\documentclass[runningheads]{llncs}
\usepackage{graphicx}
\usepackage{hyperref}
\usepackage{bbm}
\usepackage{amsmath}
\usepackage{amssymb}
\usepackage{cleveref}
\usepackage{todonotes}
\usepackage{caption}
\usepackage{subcaption}
\usepackage{booktabs}
\usepackage{multirow}
\usepackage{pifont}
\newcommand{\cmark}{\ding{51}}%
\newcommand{\xmark}{\ding{55}}%

\begin{document}
\title{On the Prediction Instability of Graph Neural Networks}
\titlerunning{Prediction Instability of GNNs}
%

\author{Max Klabunde\orcidID{0000-0002-7805-4725}\\
Florian Lemmerich\orcidID{0000-0001-7620-1376}}  
\authorrunning{M. Klabunde and F. Lemmerich}
%
\institute{
Faculty of Computer Science and Mathematics, University of Passau, Germany\\
\email{\{max.klabunde,florian.lemmerich\}@uni-passau.de}}

\maketitle              

\begin{abstract}
Instability of trained models, i.e., the dependence of individual node predictions on random factors, can affect reproducibility, reliability, and trust in machine learning systems.
In this paper, we systematically assess the prediction instability of node classification with state-of-the-art Graph Neural Networks (GNNs).
With our experiments, we establish that multiple instantiations of popular GNN models trained on the same data with the same model hyperparameters result in almost identical aggregated performance, but display substantial disagreement in the predictions for individual nodes. 
We find that up to one third of the incorrectly classified nodes differ across algorithm runs.
We identify correlations between hyperparameters, node properties, and the size of the training set with the stability of predictions. 
In general, maximizing model performance implicitly also reduces model instability.
\footnote{Code available at \href{https://github.com/mklabunde/On-the-Prediction-Instability-of-Graph-Neural-Networks}{https://github.com/mklabunde/On-the-Prediction-Instability-of-Graph-Neural-Networks}}
\end{abstract}

\section{Introduction}
Intuitively, if we fit any machine learning model with the same hyperparameters and the same data twice, we would expect to end up with the same fitted model twice.
However, recent research has found that due to random factors, such as random initializations or undetermined orderings of parallel operations on GPUs, different training runs can lead to significantly different predictions for a significant part of the (test) instances, see for example \cite{bhojanapalli_reproducibility_2021,summers_nondeterminism_2021,zhuang_randomness_2021}.
This \emph{prediction instability} (also called \emph{prediction differences} or \emph{prediction churn}) is undesirable for several reasons, including reproducibility, system reliability, and potential impact on user experience. 
For example, if a service is offered based on some classification of users with regularly retrained models, prediction instability can lead to fluctuating recommendations although there was no change of the users.
Furthermore, if only one part of a machine learning system is retrained, the subsequent parts may not be able to adapt to the difference in predictions and overall system performance deteriorates unpredictably despite improvement of the retrained model \cite{liu_model_2022}.
Finally, the reproducibility of individual predictions is important in critical domains such as finance or medicine, in which recommendations reliant on (for example) random initializations might not be acceptable.

\begin{figure}[tb]
    \centering
    \includegraphics{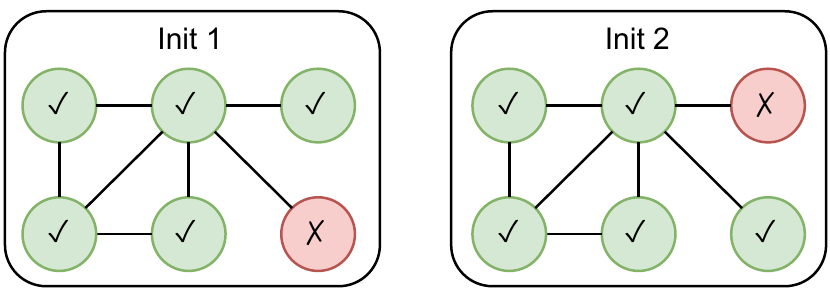}
    \caption{
    An example of prediction instability.
    Green nodes (\cmark) denote correct predictions, red ones (\xmark) false predictions.
    The two sparsely connected nodes on the right are predicted differently depending on the initialisation.
    Although the performance is identical between runs, one third of the predictions are different.
    }
    \label{fig:intro}
\end{figure}

Due to this importance, there has been a recent surge of work studying the prediction instability of machine learning models \cite{bhojanapalli_reproducibility_2021,jiang_churn_2021,liu_model_2022,milani_fard_launch_2016,shamir_anti-distillation_2020,summers_nondeterminism_2021,zhuang_randomness_2021}. 
However, research on the instability of models in graphs/network settings, such as node classification, has received little attention so far. 
As an exception, two recent studies \cite{schumacher_effects_2020,wang_towards_2020} assessed the stability of unsupervised node embedding methods, mainly from a geometrical perspective. An evaluation of state-of-the-art supervised node classification algorithms based on Graph Neural Networks (GNNs) has ---to the best of the authors' knowledge--- not yet been performed.

This paper aims to fill this research gap by presenting an extensive and systematic experimental evaluation of the prediction instability of graph neural networks.
In addition, we set out to understand how design, data, and training setup affect prediction stability.
In more detail, we summarize  the major contributions of our paper as follows: 
\begin{enumerate}
    \item We demonstrate that the popular Graph Convolutional Networks \cite{kipf_semi-supervised_2017} and Graph Attention Networks \cite{velickovic_graph_2018} exhibit significant prediction instability (\Cref{sec:model_stability}). 
    As a key result, we establish that while the aggregated accuracy of the algorithms is mostly stable, up to a third of incorrectly classified nodes differ between training runs of a model.
    
    \item We empirically study the influence of node properties (\Cref{sec:nodeprops}), model hyperparameters and the training setup on prediction instability (\cref{sec:sensitivity_training_size}).
    We find that nodes that are central in the network are less likely to be unstably predicted.
    High width, L2 regularization, low dropout rate, and low depth all show a tendency to help decrease prediction instability.
    
    \item By introspecting individual deep GNN models with centered kernel alignment \cite{kornblith_similarity_2019}, we discover a trend that deeper layers (closer to the output) are less stable (\Cref{sec:CKA}).
\end{enumerate}

In general, we find that the objectives of maximizing performance and minimizing prediction instability almost always align.
Our results have direct implications for practitioners who seek to minimize prediction instability, such as in high-stakes decision-recommendation scenarios.

\section{Preliminaries and Experimental Setup}
This section introduces our problem setting and describes the models, datasets, and instability measures used in our study.

\noindent\textbf{Multiclass Node Classification.}
We focus on the multiclass classification problem on graphs $G=(V,E)$, where every node $v \in V$ has some features $x \in \mathcal{X}$ and a one-hot encoded label $y \in \mathbf{1}^C$ with $C$ the number of different classes, and $E$ is the set of edges.
We only consider the transductive case, in which all edges and nodes including their features are known during training, but only a subset of node labels is available.

\noindent\textbf{Graph Neural Networks.}
Graph Neural Networks \cite{scarselli_graph_2009} operate by propagating information over the graph edges. 
For a specific node, the propagated information from neighboring nodes is aggregated and combined with its own representation to update its representation.
For node classification, the representations of the last layer can be used to predict the node labels.

In particular, we study in this paper two of the currently most popular state-of-the-art models for node classification: Graph Convolutional Networks (GCN) \cite{kipf_semi-supervised_2017} and Graph Attention Networks (GAT) \cite{velickovic_graph_2018}. 
GAT and GCN differ, as the aggregation mechanism of GCN uses a static normalization based on the degree of nodes, whereas GAT aggregation employs a trained multi-head attention mechanism.
We select the hyperparameters as stated in their respective papers.
The hidden dimension is 64 for GCN, GAT uses 8 attention heads with 8 dimensions each.
The models have two layers, with the second layer producing the classification output.
For the final GAT layer, the outputs of the different heads are summed up, in contrast to concatenation in the earlier layer.
In addition to the convolutional layers, we use dropout on the input and activations of the first layer with $p=0.6$.
We apply the same dropout rate to the attention weights in the GAT layer.
GCN uses ReLU activation, and GAT uses ELU. For more details on the investigated models, we refer to the original publications.

\noindent\textbf{Training Procedure.}
We train the models for a maximum of 500 epochs with an early stopping period of 40 on the validation loss using the Adam optimizer with a learning rate of 0.01.
In all cases, we use full batch training.
For each dataset, we train 50 models with different initializations.
All other sources of randomness are kept constant, with the exception of non-deterministic GPU operations.
In the following, we always present results using GPU trained models.
We go into more detail on the effect of GPU nondeterminism in \Cref{sec:discussion}.

\noindent\textbf{Datasets.}
We focus on the node classification task and use the following publicly available standard datasets:
CiteSeer and Pubmed \cite{yang_revisiting_2016}, Coauthor CS and Physics \cite{shchur_pitfalls_2019}, Amazon Photo and Computers \cite{shchur_pitfalls_2019}, and WikiCS \cite{mernyei_wiki-cs_2020}.
Since the datasets from \cite{shchur_pitfalls_2019} do not have public train/validation/test splits, we create splits for them by randomly taking 20 nodes from each class for training and using 500 nodes as validation data. 
The rest are used for testing.
All datasets are treated as undirected graphs.

\noindent\textbf{Measuring Prediction Instability.}
\label{sec:measuring}
To quantify prediction instability, we now define several measures that capture differences or disagreements in the model output.
These measures have two main differences:
They either use the predicted labels as input, i.e., the argmax of the model output, or the softmax-normalized output.
While the first approach directly follows intuition and has the advantage of being very interpretable, it has the downside of using the discontinuous argmax, which means that even slight differences in model output can lead to different outcomes.

We follow the definition of Madani et al. \cite{madani_co-validation_2004} and define the (expected) prediction disagreement as follows:
\begin{equation}
    d = \mathbb{E}_{x, f_1, f_2} \mathbbm{1}\{\arg \max f_1(x) \neq \arg \max f_2(x)\}
\end{equation}
where $f_i \in F$ are instantiations of a model family $F$ and $\mathbbm{1}\{\cdot\}$ is the indicator function.
The disagreement is easily calculated in practice by training a number of models and then averaging the pairwise disagreements.
This measure is also known as churn \cite{bahri_locally_2021,bhojanapalli_reproducibility_2021,jiang_churn_2021,milani_fard_launch_2016} and jitter \cite{liu_model_2022}.

The theoretically possible value of the disagreement of a pair of model instantiations is bounded by their performance.
For example, when two models, $f_1$ and $f_2$, perform with 95\% accuracy, then the minimal disagreement equals zero, which occurs when the predictions are identical.
Maximal disagreement occurs when 90\% of predictions are identical, and $f_1$ is correct on the 5\% of data where $f_2$ is incorrect.
In general, it holds \cite{bhojanapalli_reproducibility_2021}:
\begin{equation}
    |Err_{f_1} - Err_{f_2}| \leq d_{f_1, f_2} \leq \min(1, Err_{f_1} + Err_{f_2})
\end{equation} where $Err$ is the error rate of a model and $d_{f_1, f_2}$ is the empirical disagreement between $f_1$ and $f_2$.
While two models with high error rates do not necessarily have large disagreement, we later show that disagreement and error rate are in fact highly correlated, a finding which has not received much attention so far.

Absolute disagreement, as defined previously, is a very intuitive and straightforward measure of disagreement.
However, to better understand the relationship between disagreement and error rate, we define the min-max normalized disagreement, which gives the disagreement relative to its minimal and maximal possible values:
\begin{equation}
    d_{norm} = \mathbb{E}_{f_1, f_2}\left[ \frac{d_{f_1, f_2} - \min d_{f_1, f_2}}{\max d_{f_1, f_2} - \min d_{f_1, f_2}}\right].
\end{equation}

A natural extension of the aforementioned measures is to condition the computation on specific subgroups of predictions, e.g., the correct or incorrect predictions.
As Milani Fard et al. \cite{milani_fard_launch_2016}, we define those as true disagreement $d_{True}$ and false disagreement $d_{False}$, respectively:
\begin{equation}
    d_{True} = \mathbb{E}_{(x,y), f_1, f_2} \mathbbm{1}\{\arg \max f_1(x) \neq \arg \max f_2(x)| \arg \max f_1(x) = y\}.
\end{equation}
$d_{False}$ is computed analogously.
False disagreement gives an indication of prediction stability irrespective of model performance as it is always possible that incorrectly predicted nodes are predicted differently in another run.
False disagreement and normalized disagreement are especially important to disentangle model performance and stability since model hyperparameters, such as width, may affect performance and disagreement jointly.

All of the above measures work on the hard predictions of the models.
To measure the difference between the output distributions, we use the simple mean absolute error, where $C$ is the number of classes:
\begin{align}
    d_{MAE} &= \mathbb{E}_{x, f_1, f_2}\left[ \frac{1}{C} ||f_1(x) - f_2(x)||_1\right].
\end{align}

\section{Results}
This section presents our main results on the instability of GNNs. 
We showcase overall results before introducing detailed analyses on the effect of node properties and model design.
Finally, we describe results of inspecting the instability of GNNs layer-by-layer.
We always report results on examples not seen during training.

\subsection{Overall Prediction Instability of GNNs}
\label{sec:model_stability}

We now demonstrate the prediction instability of GAT and GCN on several well-known datasets.

\begin{table}[tb]
\centering
\scriptsize
\caption{Prediction disagreement (in \%) and its standard deviation on different datasets.}
\label{tab:baseline}
\setlength{\tabcolsep}{2pt}
\begin{tabular}{llcccccc}
\toprule
Dataset & Model &       Accuracy &               $d$ &        $d_{norm}$ &       $d_{True}$ &       $d_{False}$ &              MAE \\
\midrule
\multirow{2}{*}{CiteSeer} & GAT &  69.04 $\pm$ 0.88 &  10.31 $\pm$ 1.70 &  15.30 $\pm$ 2.61 &  5.11 $\pm$ 1.28 &  21.83 $\pm$ 3.76 &  3.32 $\pm$ 0.53 \\
       & GCN &  69.11 $\pm$ 0.61 &   7.32 $\pm$ 1.04 &  10.85 $\pm$ 1.77 &  3.58 $\pm$ 0.86 &  15.66 $\pm$ 2.50 &  2.97 $\pm$ 0.36 \\
\cline{1-8}
\multirow{2}{*}{Pubmed} & GAT &  75.69 $\pm$ 0.60 &   3.75 $\pm$ 1.30 &   6.41 $\pm$ 2.62 &  2.38 $\pm$ 1.00 &   7.95 $\pm$ 3.18 &  2.25 $\pm$ 0.69 \\
       & GCN &  76.78 $\pm$ 0.55 &   2.57 $\pm$ 0.78 &   4.25 $\pm$ 1.50 &  1.57 $\pm$ 0.70 &   5.84 $\pm$ 2.32 &  2.49 $\pm$ 0.94 \\
\cline{1-8}
\multirow{2}{*}{CS} & GAT &  90.70 $\pm$ 0.43 &   3.64 $\pm$ 0.46 &  17.52 $\pm$ 1.91 &  1.74 $\pm$ 0.40 &  22.04 $\pm$ 3.33 &  0.68 $\pm$ 0.10 \\
       & GCN &  90.80 $\pm$ 0.44 &   3.28 $\pm$ 0.59 &  15.51 $\pm$ 2.64 &  1.57 $\pm$ 0.46 &  20.02 $\pm$ 4.08 &  0.67 $\pm$ 0.21 \\
\cline{1-8}
\multirow{2}{*}{Physics} & GAT &  91.89 $\pm$ 0.67 &   3.84 $\pm$ 0.78 &  19.87 $\pm$ 4.07 &  1.86 $\pm$ 0.64 &  25.93 $\pm$ 6.21 &  2.00 $\pm$ 0.52 \\
       & GCN &  92.67 $\pm$ 0.34 &   1.64 $\pm$ 0.42 &   8.76 $\pm$ 2.69 &  0.79 $\pm$ 0.33 &  12.28 $\pm$ 4.22 &  1.29 $\pm$ 0.45 \\
\cline{1-8}
\multirow{2}{*}{Computers} & GAT &  80.85 $\pm$ 1.39 &   8.99 $\pm$ 2.27 &  20.31 $\pm$ 5.48 &  4.47 $\pm$ 1.75 &  27.92 $\pm$ 7.37 &  2.21 $\pm$ 0.50 \\
       & GCN &  80.96 $\pm$ 1.21 &  10.50 $\pm$ 2.34 &  24.94 $\pm$ 5.28 &  5.08 $\pm$ 1.61 &  33.34 $\pm$ 6.93 &  2.43 $\pm$ 0.52 \\
\cline{1-8}
\multirow{2}{*}{Photo} & GAT &  90.39 $\pm$ 0.62 &   3.93 $\pm$ 0.86 &  17.40 $\pm$ 4.02 &  1.81 $\pm$ 0.64 &  23.68 $\pm$ 5.80 &  1.42 $\pm$ 0.30 \\
       & GCN &  90.76 $\pm$ 0.50 &   3.84 $\pm$ 0.82 &  18.29 $\pm$ 4.04 &  1.70 $\pm$ 0.54 &  24.69 $\pm$ 5.49 &  1.47 $\pm$ 0.26 \\
\cline{1-8}
\multirow{2}{*}{WikiCS} & GAT &  79.58 $\pm$ 0.23 &   3.72 $\pm$ 0.52 &   8.54 $\pm$ 1.24 &  1.72 $\pm$ 0.32 &  11.54 $\pm$ 1.75 &  0.90 $\pm$ 0.14 \\
       & GCN &  79.42 $\pm$ 0.20 &   3.27 $\pm$ 0.42 &   7.43 $\pm$ 1.01 &  1.54 $\pm$ 0.27 &   9.91 $\pm$ 1.40 &  0.73 $\pm$ 0.09 \\
\bottomrule
\end{tabular}

\end{table}

We show the results in \Cref{tab:baseline}.
The prediction disagreement $d$ is between three and four percent, with the exceptions of CiteSeer and Computers, where the disagreement is over ten percent.
Models with higher accuracy tend to have lower disagreement, but the datasets must be taken into account.
For example, the classification accuracies are higher on the Computers dataset compared to Pubmed, but disagreement is higher as well.
We find that $d_{norm}$ is lower than 25 percent, which means that disagreement is relatively close to the minimum that variation in model performance allows.
Interestingly, nodes that are falsely predicted by one model have a high probability of being predicted differently by another model.
False disagreement is almost always at least one magnitude larger than true disagreement, which is partially explained by the high performance of the models.
Finally, the mean absolute error reveals that the predicted probabilities of classes are not much different between models. 
Overall, GNNs clearly demonstrate prediction instability to a significant degree.

In the following, we focus in our discussion on the prediction disagreement measure $d$ as the most indicator. Typically, the same tendencies can be observed for the other measures.\footnote{See the supplementary material for explicit results.}

\subsection{The Effect of Node Properties}
\label{sec:nodeprops}
Next, we examine the unveiled prediction instability in more detail and set them in relation to data properties.
We consider four node properties: i) PageRank, ii) the clustering coefficient, iii) the $k$-core, and iv) the class label.
The first three are related to the graph structure, whereas the class label is related to underlying node features.
PageRank measures the centrality of a node, the clustering coefficient the connectivity of a node neighborhood, and the $k$-core of a node is the maximal $k$, for which the node is part of a maximal subgraph containing only nodes with a degree of at least $k$.
Thus, the $k$-core gives an indication of both connectivity and centrality.

For the structural properties, we divide the nodes into seven equal-sized parts (septiles) with respect to the analyzed property.
Then, we record prediction instability and model performance for each of these subgroups of the data.
We use the same models as in the previous section.

\begin{figure}[t]
    \centering
    \begin{subfigure}{0.33\textwidth}
        \includegraphics[width=\textwidth]{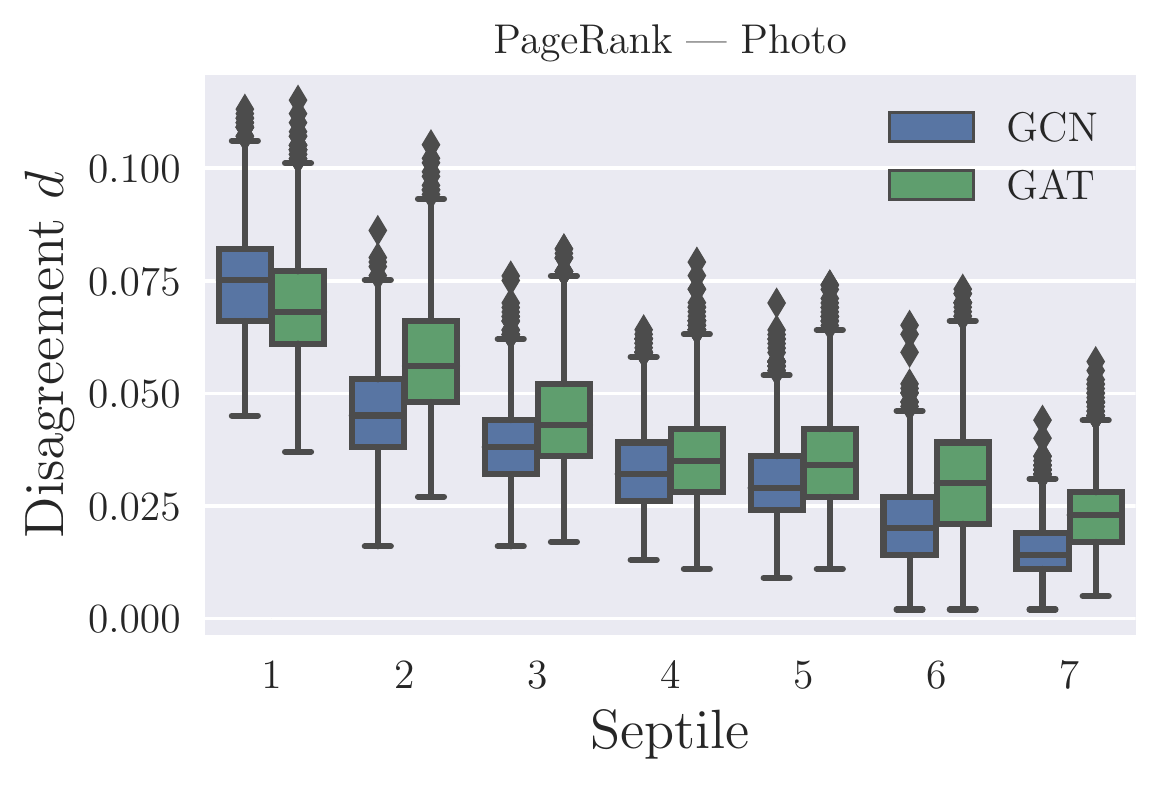}
    \end{subfigure}\hfill
    \begin{subfigure}{0.33\textwidth}
        \includegraphics[width=\textwidth]{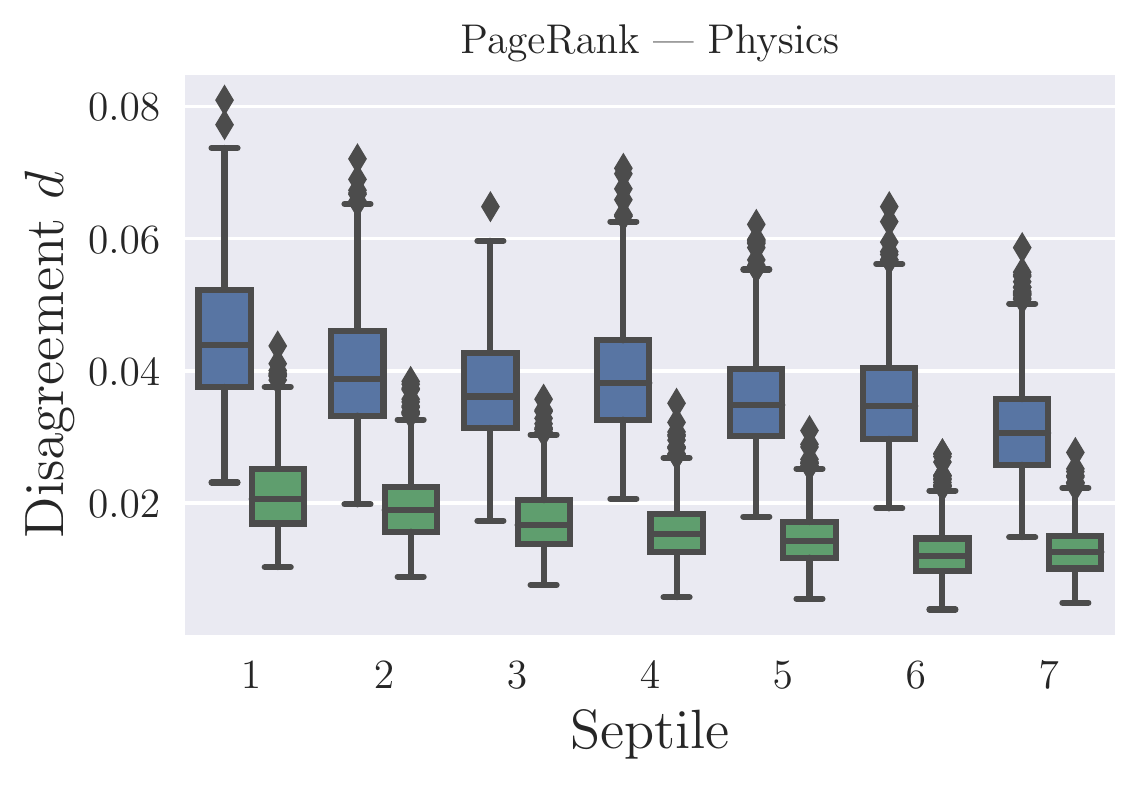}
    \end{subfigure}\hfill
    \begin{subfigure}{0.33\textwidth}
        \includegraphics[width=\textwidth]{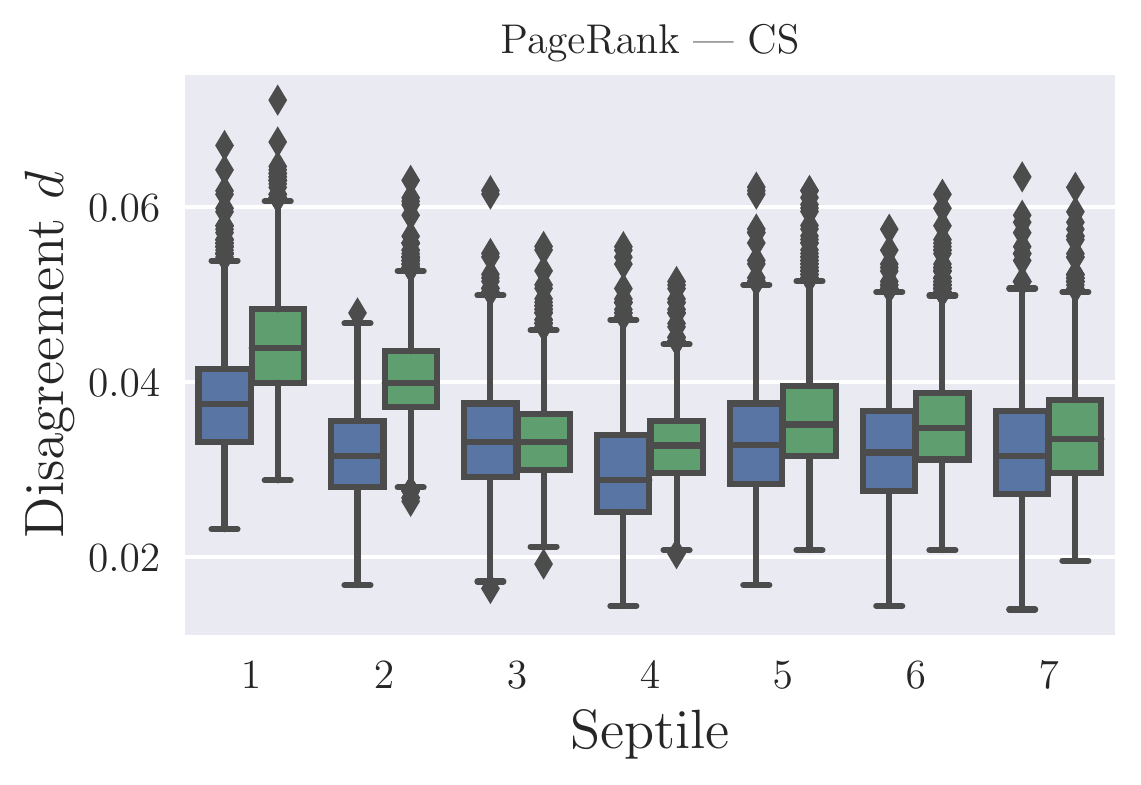}
    \end{subfigure}\vfill
    \caption{
        Prediction disagreement for PageRank septiles on selected datasets.
    }
    \label{fig:nodeprops}
\end{figure}

\begin{figure}[t]
    \centering
    \begin{subfigure}{0.5\textwidth}
        \includegraphics[width=\textwidth]{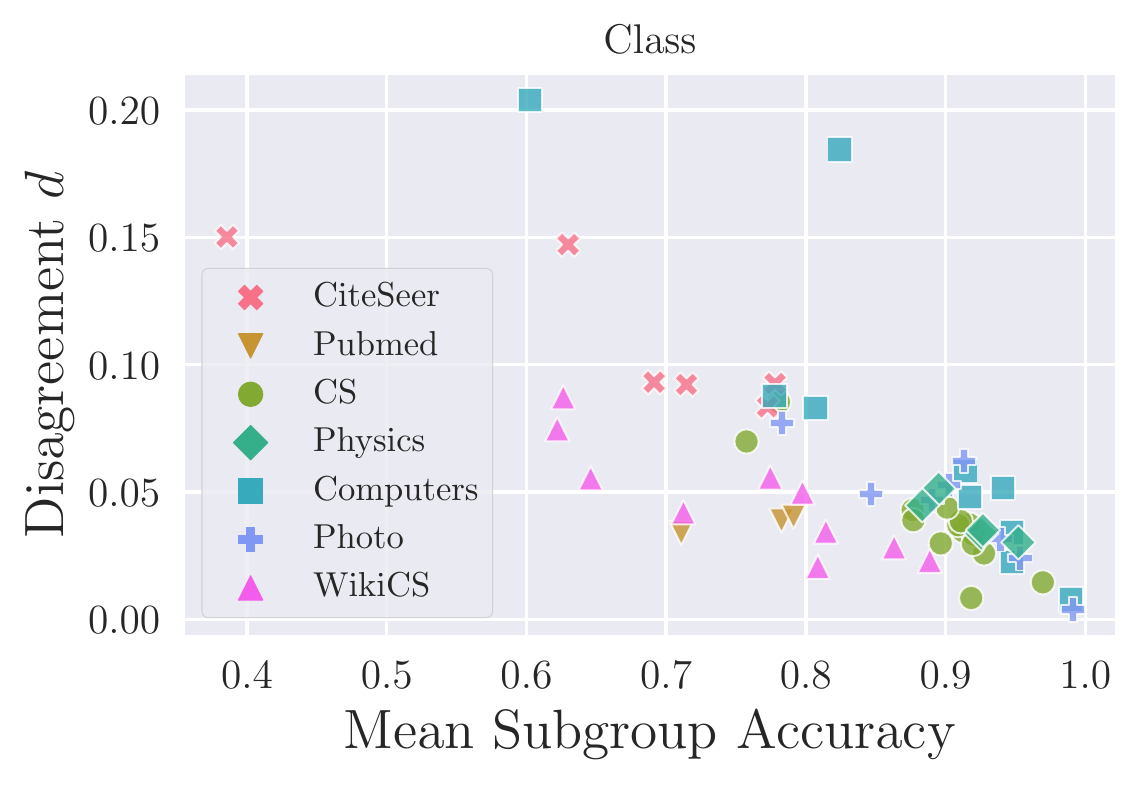}
    \end{subfigure}\hfill
    \begin{subfigure}{0.5\textwidth}
        \includegraphics[width=\textwidth]{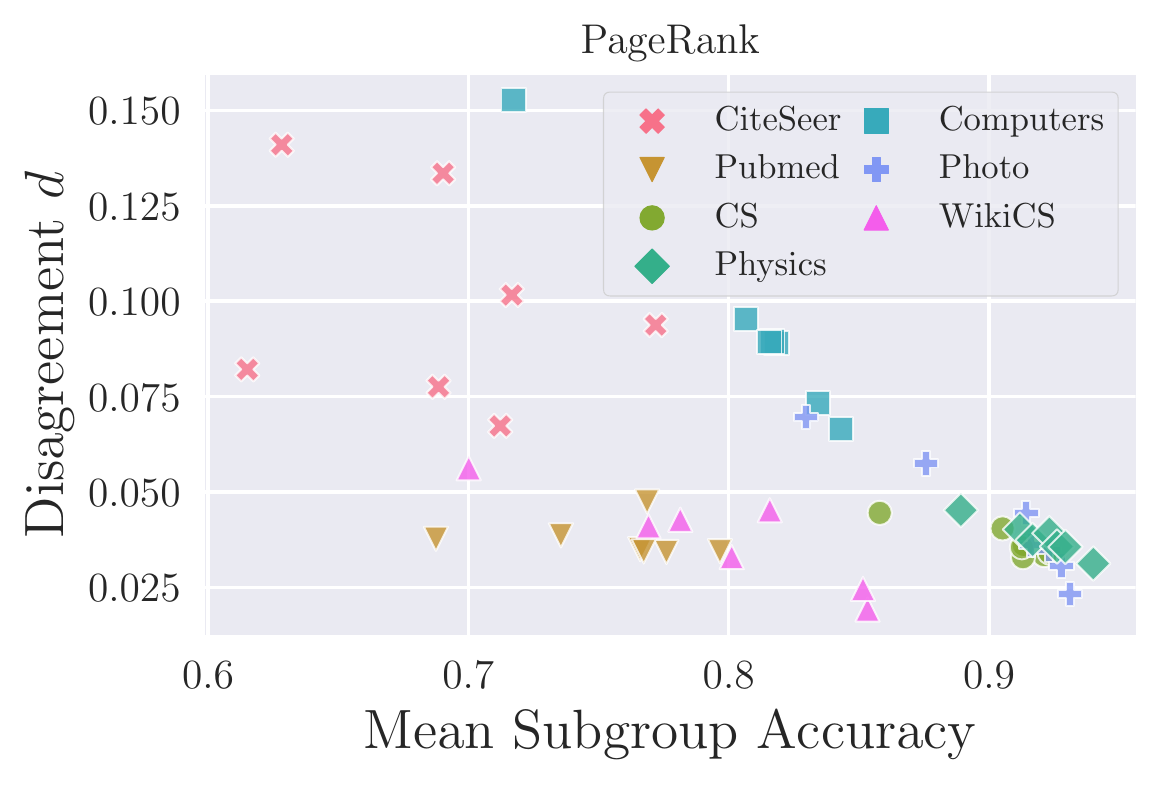}
    \end{subfigure}
    \caption{
        Relation between subgroup accuracy and prediction disagreement for GAT with respect to the node classes (left) and PageRank septiles (right).
        The higher the accuracy, the lower the disagreement.
        The correlation between accuracy and disagreement suggests that central nodes are stably predicted due to high model performance on that subgroup of the data.
        Results for the other properties and GCN are similar.
    }
    \label{fig:subgroupacc_to_pi}
\end{figure}

We give a representative overview of prediction disagreement in relation to the values of structural properties in \Cref{fig:nodeprops}.
Central nodes show lower prediction instability.
Low clustering negatively impacts stability, but there is no consistent relation over all datasets for other septiles.
A high core number reduces the risk of prediction instability.
For all properties, the results depend on the dataset to some extent.
We attribute many of the differences in prediction instability to differences in model performance in different subgroups (\Cref{fig:subgroupacc_to_pi}).
Now, we describe the results in more detail.

\noindent\textbf{Structural Properties.}
Nodes with higher PageRank have lower prediction disagreement on all datasets except Pubmed.
However, the magnitude of the difference varies depending on the dataset.
Interestingly, the disagreement of false predicted nodes is roughly constant in many cases.
This also holds for normalized disagreement, which shows that differences between subgroups can be explained by differences in accuracy.
The MAE of the output distributions almost always decreases with higher PageRank.

There is no consistent relationship between the clustering coefficient and the prediction instability in our results.
The only common trend is that low clustering is correlated with high prediction disagreement.
Higher clustering coincides with lower prediction disagreement on WikiCS and CS, Physics and Computers display an U-like relationship.
On Photo, there is no clear trend.
For Pubmed and CiteSeer, more than 65\% of the nodes have a clustering coefficient of zero, so a comparison of equal bins is not possible.
Similar to PageRank, false and normalized disagreement are almost constant in many cases, which again highlights that the prediction disagreement is closely related to the model performance.

On all datasets, the group with the highest k-core has the lowest prediction disagreement, and the group with the lowest k-core has the highest disagreement.
In between, we do not observe a clear trend.
Furthermore, the variance of the disagreement decreases with increasing k-core.
Again, false disagreement and normalized disagreement do not show large differences.

\noindent\textbf{Class Label.}
There are large differences between classes with respect to both the average disagreement and its variance.
Differences shrink for false disagreement and when normalizing for accuracy.
The relation of MAE to class labels mirrors that of prediction disagreement.

Classes with few examples are not less stable than large classes.
As shown in \Cref{fig:subgroupacc_to_pi}, the average accuracy explains much of the differences in disagreement.
Interestingly, the variance of the accuracy does not impact the prediction disagreement, although it affects the lower bound of prediction disagreement, which depends on the performance difference of two models.

\subsection{The Effect of Model Design and Training Setup}
\label{sec:sensitivity_training_size}
In the previous section, we find evidence that prediction stability is related to model performance.
Model design and training setup reasonably influence model performance (and thus may impact prediction stability), but how exactly they correlate with prediction stability and if they influence stability beyond the performance is unclear.
We test the influence of individual hyperparameters on prediction stability by following the training protocol of \Cref{sec:measuring}, but changing one hyperparameter per experiment, if not specified otherwise.

\noindent\textbf{Training Data. }
\begin{figure}[tb]
    \centering
    \begin{subfigure}[b]{0.5\textwidth}
        \includegraphics{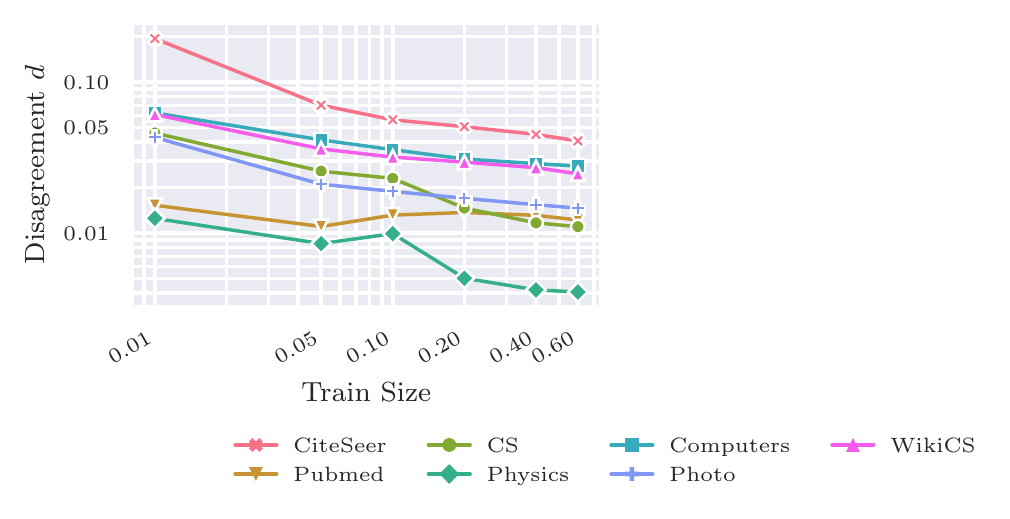}
    \end{subfigure}\hfill
    \begin{subfigure}[b]{0.5\textwidth}
        \includegraphics{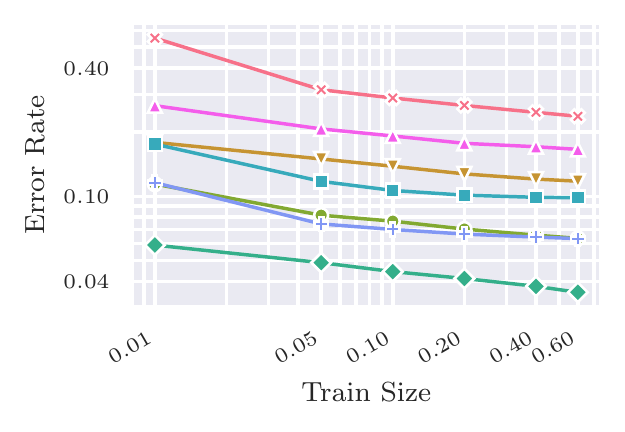}
        \vspace{14pt}
    \end{subfigure}
    \caption{
    Effect of amount of training data on disagreement (left) and error rate (right) for GCN.
    }
    \label{fig:trainingdata}
\end{figure}
To analyze the effect of training data availability on prediction stability, we vary the number of node labels available for training between 1 and 60\% of all nodes (steps: 0.01, 0.05, 0.1, 0.2, 0.4, 0.6) and use a fixed-size validation set of 15\% of the data.
We sample the nodes of each class proportionally to their total class size, so that nodes of all classes are present in the test set.
Further, to avoid dependency on specific data splits, we repeat the experiment with 10 different data splits per graph.
In total, we train 42000 models for this experiment.

Results are shown on a log-log scale in \Cref{fig:trainingdata}. We find that both the disagreement and the error rate decrease significantly with increasing available training data. This underlines the correlation between model performance and disagreement. The same trends can also be observed also for the other measures of disagreement.
Only Pubmed shows significantly different behaviour; that is, the disagreement does not decrease.
GAT results are highly similar and also show a smooth decrease in disagreement and error rate.

\noindent\textbf{Optimizer. }
\begin{figure}[tb]
    \centering
    \begin{subfigure}[b]{0.5\textwidth}
        \includegraphics{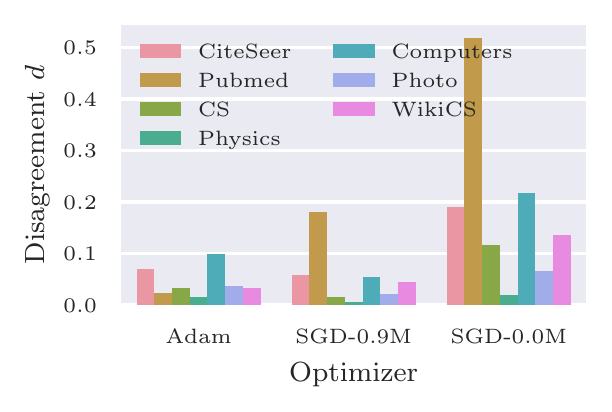}
    \end{subfigure}\hfill
    \begin{subfigure}[b]{0.5\textwidth}
        \includegraphics{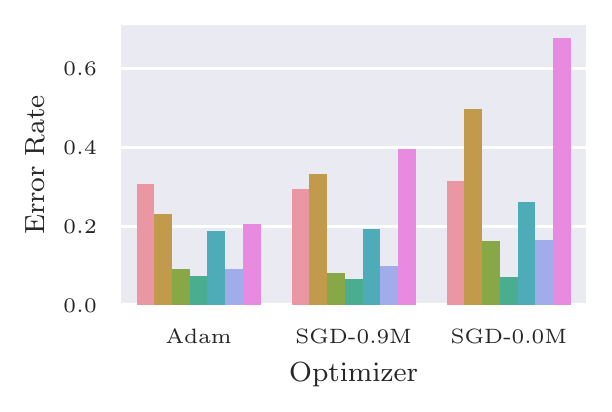}
    \end{subfigure}
    \caption{
    Prediction disagreement (left) and error rate (right) for training GCN with Adam, SGD with momentum of 0.9, and SGD without momentum. 
    }
    \label{fig:optim}
\end{figure}
We train the models with different optimizers: Adam, Stochastic Gradient Descent (SGD), and SGD with momentum (SGD-M), which we set to 0.9.
We show the results for GCN in \Cref{fig:optim}.
For prediction disagreement, SGD performs much worse than SGD-M and Adam.
SGD-M performs on par with or better than Adam, with the exception of Pubmed.
Overall, disagreement and error rate are correlated, but SGD-M optimization leads to lower disagreement.
Furthermore, SGD-M decreases the average MAE between the output distributions of the models more than Adam.
For GAT, Adam and SGD-M perform similarly with a slight edge to Adam.
SGD performs much worse with respect to both error rate and disagreement.
Based on the discrepancy between GCN and GAT results, there does not appear to be a simple rule to select one of the tested optimizers to generally minimize prediction disagreement.

\noindent\textbf{L2 Regularization. }
\begin{figure}[tb]
    \centering
    \begin{subfigure}[b]{0.5\textwidth}
        \includegraphics{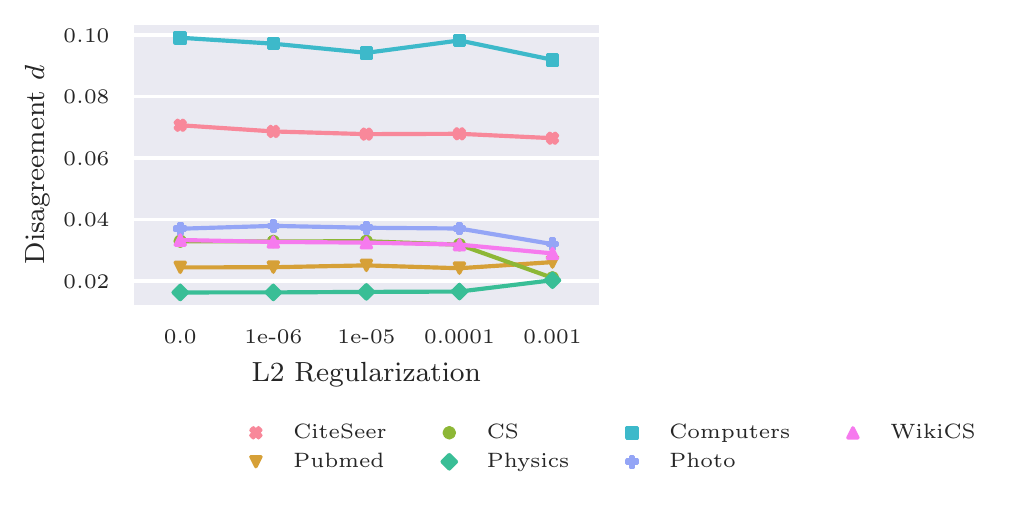}
    \end{subfigure}\hfill
    \begin{subfigure}[b]{0.5\textwidth}
        \includegraphics[width=\textwidth]{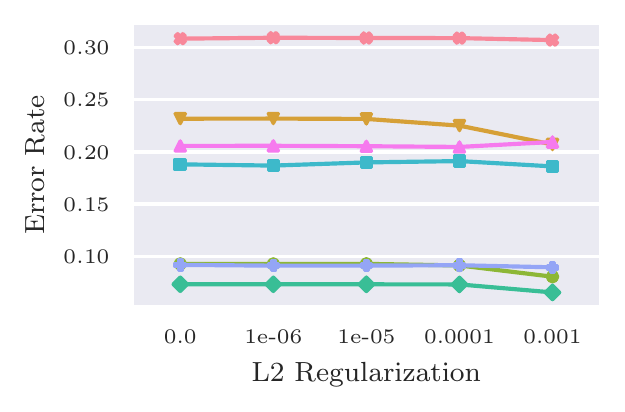}
        \vspace{16pt}
    \end{subfigure}
    \caption{
    Effect of L2 regularization on disagreement (left) and error rate (right) for GCN. 
    }
    \label{fig:l2}
\end{figure}
We show the results of GCN for varying L2 regularization parameters in \Cref{fig:l2}.
Disagreement decreases only slightly with moderate regularization.
We observe the largest changes, both for the disagreement and error rate, with the maximal value of L2 regularization.
Strong regularization reduces disagreement (all measures) on 5 out of 7 datasets compared to without regularization.
Interestingly, disagreement decreases even when the error rate stays roughly constant.

\noindent\textbf{Dropout. }
\begin{figure}[tb]
    \centering
    \begin{subfigure}[b]{0.48\textwidth}
        \includegraphics{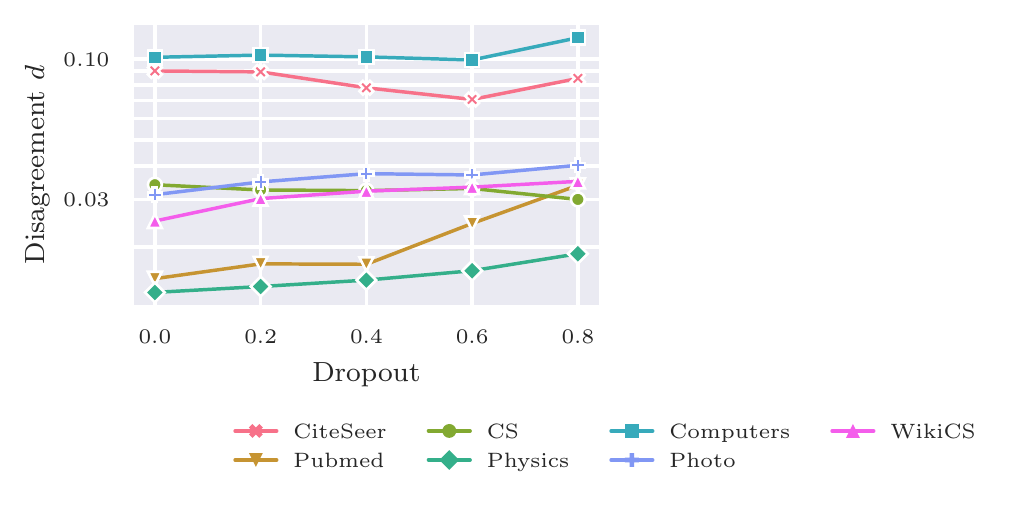}
    \end{subfigure}
    \begin{subfigure}[b]{0.5\textwidth}
        \includegraphics{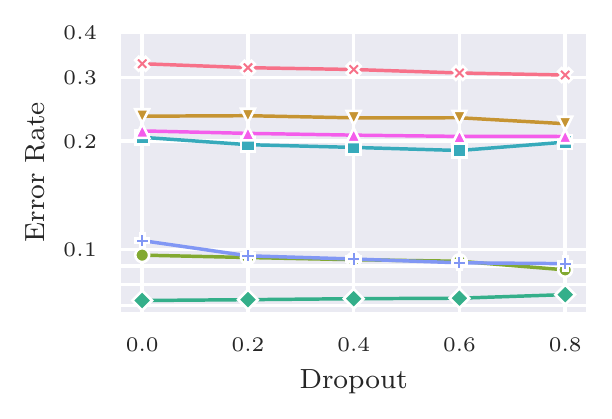}
        \vspace{15pt}
    \end{subfigure}
    \caption{
    Effect of dropout on disagreement (left) and error rate (right) for GCN.
    }
    \label{fig:dropout}
\end{figure}
We show the results for varying dropout rates in \Cref{fig:dropout}.
Large dropout rates increase disagreement for 6 of the 7 datasets for GCN.
In contrast to previous observations, change in disagreement does not follow the error rate.
Instead, the more dropout, the more prediction disagreement in most cases.
Although the effect is small in absolute terms, dropout influences prediction stability negatively.
Even so, a finely tuned dropout rate can improve disagreement in some cases while also decreasing the error rate.

\noindent\textbf{Width. }
\begin{figure}[tb]
    \centering
    \begin{subfigure}[b]{0.48\textwidth}
        \includegraphics{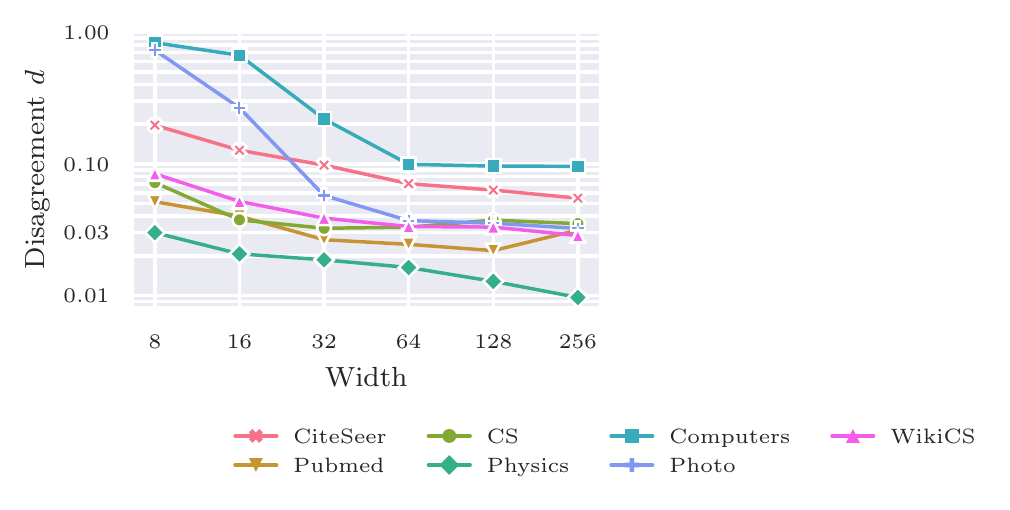}
    \end{subfigure}\hfill
    \begin{subfigure}[b]{0.5\textwidth}
        \includegraphics{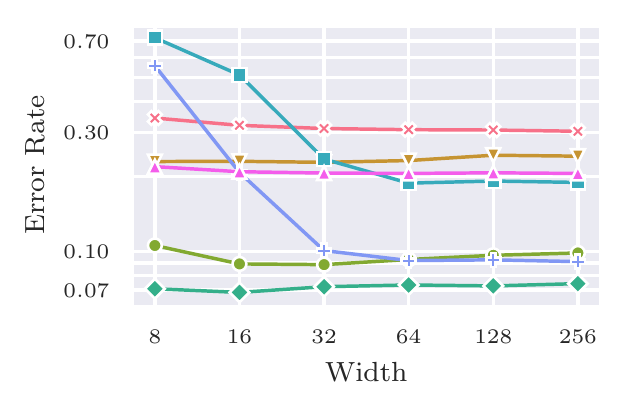}
        \vspace{15pt}
    \end{subfigure}
    \caption{
    Effect of width on disagreement (left) and error rate (right) for GCN.
    }
    \label{fig:width}
\end{figure}
We vary the width of the models on a logarithmic scale between 8 and 256.
GAT always has eight attention heads, which means that the dimension per head varies between 1 and 32.
\Cref{fig:width} shows the absolute disagreement $d$ and the error rate in relation to the width for GCN.
Wider models have less prediction disagreement, which holds even for models for which the error rate does not decrease.
This relation is mirrored in all stability measures.

\noindent\textbf{Depth. }
\begin{figure}[tb]
    \centering
    \begin{subfigure}[b]{0.5\textwidth}
        \includegraphics{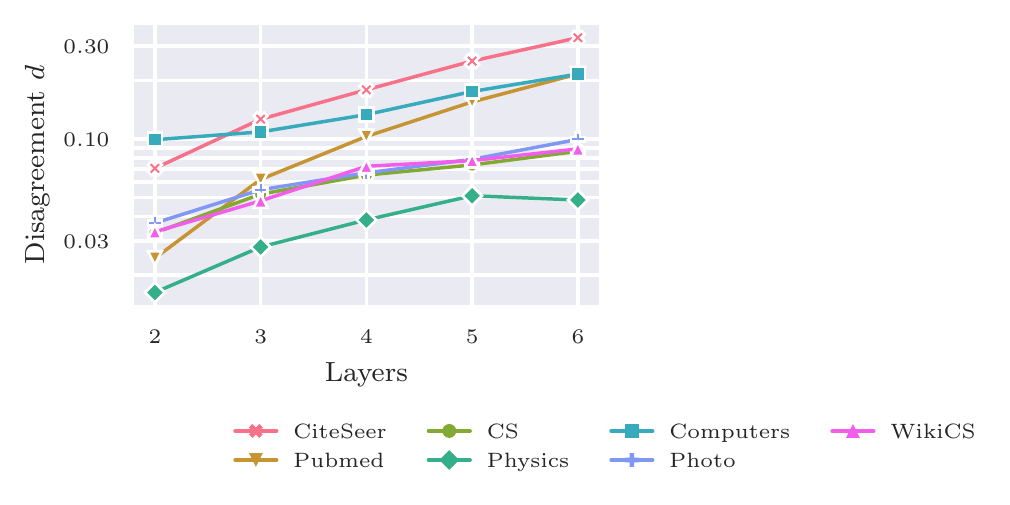}
    \end{subfigure}\hfill
    \begin{subfigure}[b]{0.5\textwidth}
        \includegraphics{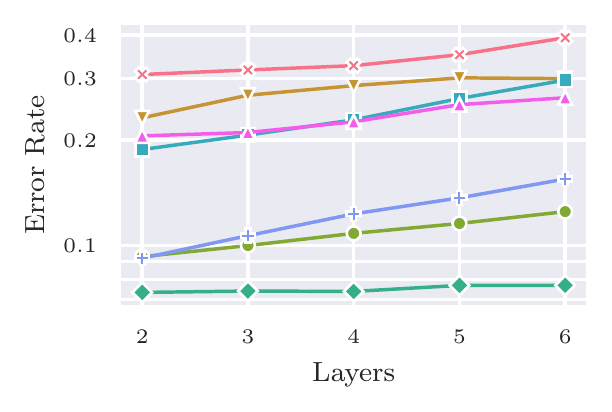}
        \vspace{15pt}
    \end{subfigure}
    \caption{
    Effect of model depth on disagreement (left) and error rate (right) for GCN.
    }
    \label{fig:depth}
\end{figure}
We change the number of layers from 2 to 6. 
Between every layer, there are dropout and activation functions, while otherwise following the previously used training procedure.
\Cref{fig:depth} shows the results for GCN.
Prediction disagreement increases with depth of the model.
Similarly, the error rate increases, which can be explained by a lack of training techniques for deep GNNs, e.g., residual connections or normalization.
Nevertheless, even when the model performance does not decrease much, prediction stability decreases, e.g., on the Physics dataset, the absolute disagreement increases almost four-fold.
We make the same observation for GAT, which suggests that the depth of a model negatively affects its prediction stability.

\noindent\textbf{Combining Optimal Hyperparameters. }
\begin{figure}[tb]
    \centering
    \begin{subfigure}[b]{0.5\textwidth}
        \includegraphics{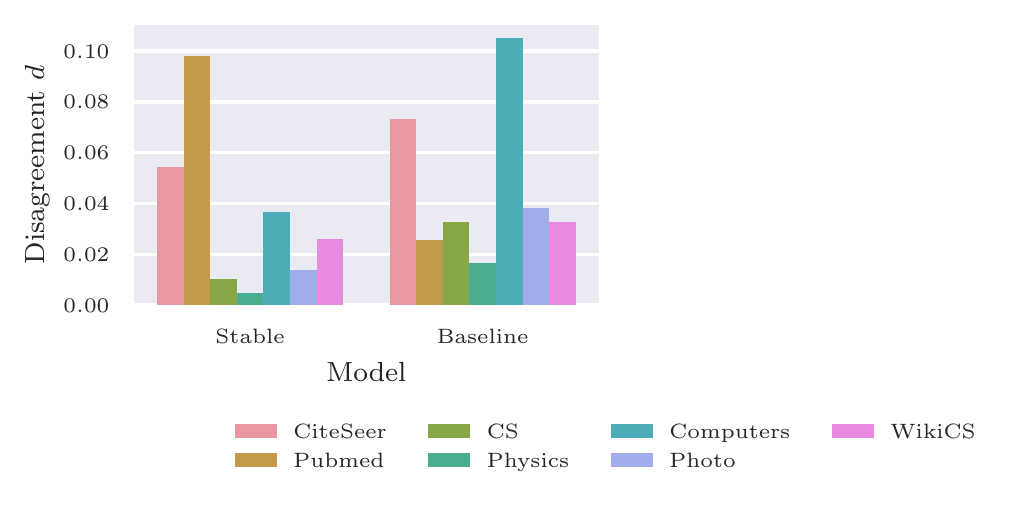}
    \end{subfigure}\hfill
    \begin{subfigure}[b]{0.5\textwidth}
        \includegraphics{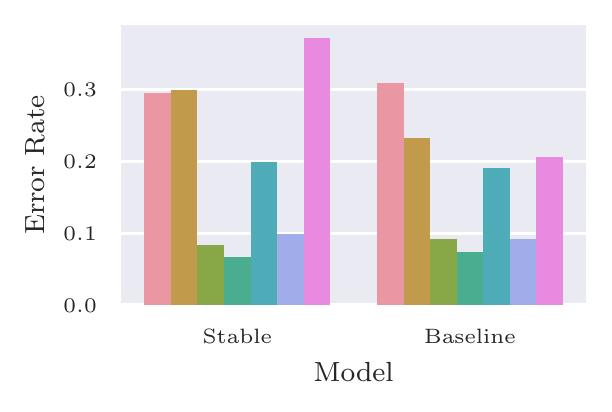}
        \vspace{15pt}
    \end{subfigure}
    \caption{
    Comparison of baseline GCN model and a stable variant, which implements all hyperparameters as suggested by the previous experiments.
    }
    \label{fig:stable}
\end{figure}
To test whether the observations so far can inform model selection, we now train ``stable variants" of GAT and GCN and compare them with the baseline models, as described in \Cref{sec:measuring}.
We select hyperparameters of the models according to the previous experiments, i.e., those that minimize the disagreement.
Since the best hyperparameters differ between datasets, we manually pick them as width of 256 (200 for GAT on Physics due to memory limitations), depth of 2, dropout of 0.2, and L2 regularization of $10^{-4}$.
We use SGD with a momentum of 0.9 and the same data as in all previous experiments.

We show the results for GCN in \Cref{fig:stable}.
The stable variant has less prediction disagreement on 6 of the 7 datasets, despite not always having a lower error rate.
We make a similar observation for GAT, having lower disagreement on 5 of the 7 datasets.
These results could be explained by the fact that the stable variants have a lower variation in model performance, which is a source of prediction instability.

\subsection{Layer-wise Model Introspection}
\label{sec:CKA}
In the last part of our analysis, we aim to obtain a better understanding about where in the deep neural architecture instability primarily arises. For that purpose, we investigate the  (in-)stability of internal representations with centered kernel alignment (CKA) \cite{kornblith_similarity_2019} to measure the similarity of representations from corresponding layers in different models, see below for a slightly more extensive description.
That is, we compare layer 1 of model A with layer 1 of model B, layer 2 of A with layer 2 of B, etc.
We focus on the similarity of models with varying depths and train the models according to \Cref{sec:measuring}, but again vary the number of layers and add dropout layers between them.

\noindent\textbf{Centered Kernel Alignment. }
CKA is a state-of-the-art method for measuring the similarity of neural network representations.
Roughly speaking, CKA compares two matrices of pairwise similarities by vectorizing them and calculating the dot product.
We use the linear variant of CKA, i.e., the pairwise similarities are calculated with the dot product, since it is efficient and other variants, such as using the RBF kernel for similarity computation, do not show better performance consistently. For details, we refer to the original publication~\cite{kornblith_similarity_2019}.

\noindent\textbf{Results. }
\begin{figure}[tb]
    \centering
    \begin{subfigure}[b]{0.5\textwidth}
        \includegraphics[width=\textwidth]{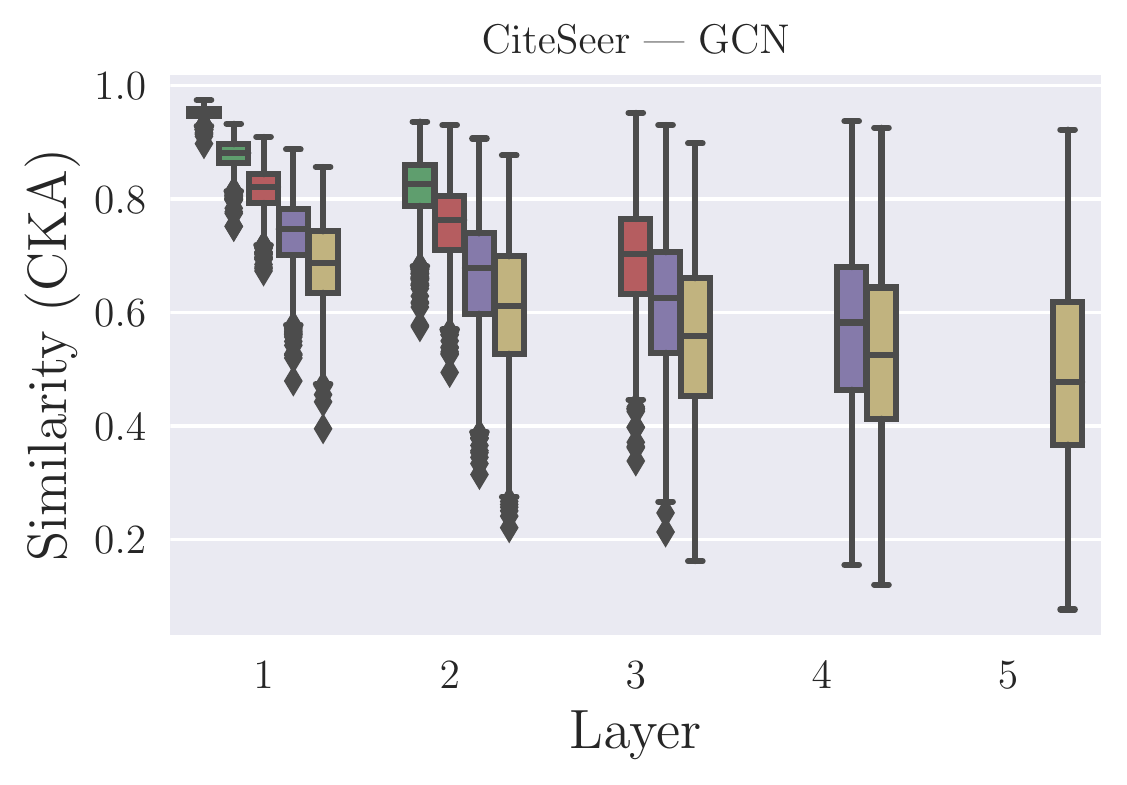}
    \end{subfigure}\hfill
    \begin{subfigure}[b]{0.5\textwidth}
        \includegraphics[width=\textwidth]{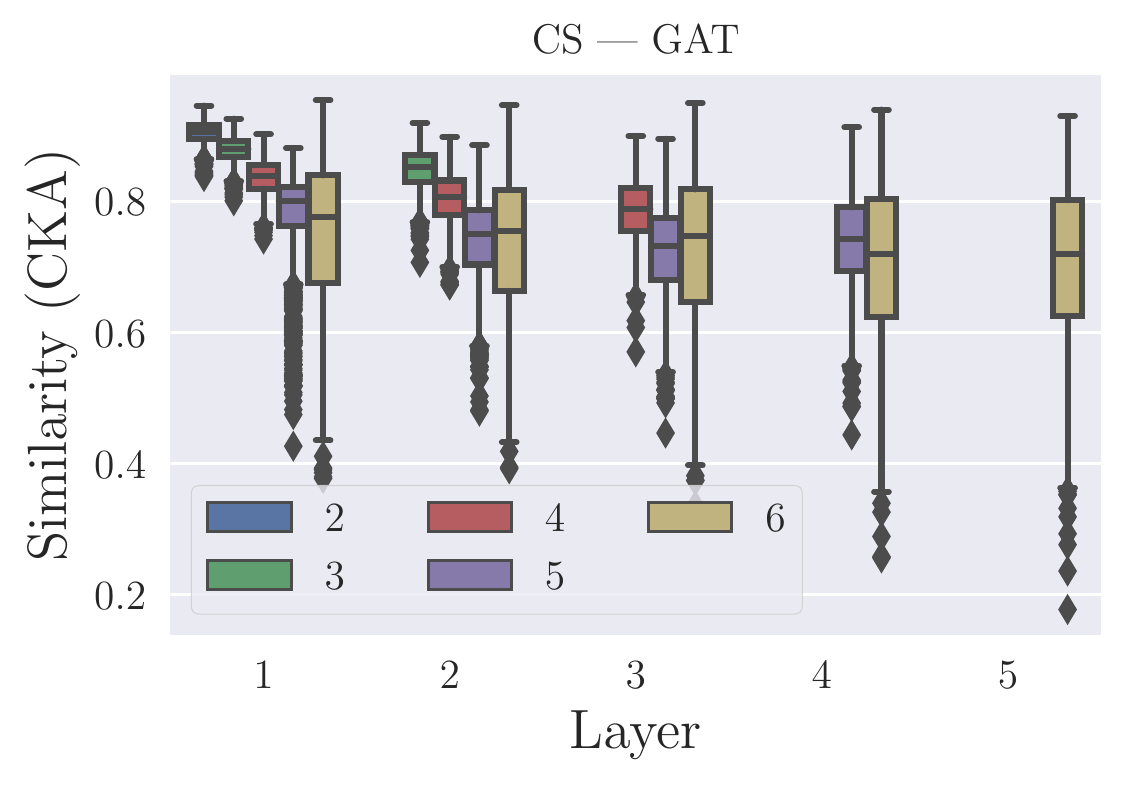}
    \end{subfigure}
    \caption{
    Similarity of layers to corresponding layers of another trained GCN on CiteSeer (left) and GAT on CS (right).
    The deeper the layer in the trained model, the less similar they are.
    }
    \label{fig:cka}
\end{figure}
\Cref{fig:cka} shows some exemplary results. In these plots, each color of boxplots refers to one model with a specific number of layers while the different groups of boxplots from left to right refer to the position of the layer within the model architecture.
In general, the more layers the model has, the lower the similarity of the layers .
Moreover, the deeper the layer (closer to the output), the lower the similarity.
Similarly, the variance of the similarity increases with depth.
Some outliers exist, for which changes in similarity between layers are small, or the similarity increases with depth (GCN WikiCS and Computers, 6 layer GAT on Pubmed).
Overall, however, deep GNNs have more variance in their internal structure and representations.

Although the first layers may suffer from vanishing gradients, they are much more self-similar than the deeper layers, which should receive much larger updates.
On the one hand, these large updates could make deep layers less similar as they may need to adapt to varying outputs of the earlier layers.
On the other hand, the first layers are extremely similar, although they start from a random initialization.
As a consequence, these layers provide very similar representations to deep layers, questioning why the deep layers are dissimilar.
We leave a more detailed analysis for future work and hope that this observation sparks further research into the learned representations of GNNs.

\section{Discussion}
\label{sec:discussion}
We discuss limitations and implications of our work, as well as avenues for future work.

\noindent\textbf{Limitations.}
The models we study do not use popular techniques for deep GNNs, such as normalization or residual connections.
Furthermore, we avoid mini-batching and distributed training.
Although relatively shallow GNNs work well on many tasks, recent work introduces new benchmarks that benefit greatly from more complex models \cite{dwivedi_benchmarking_2020,hu_ogb-lsc_nodate}.
Therefore, interesting future work would be to explore how these techniques, combined with larger models and larger graphs, affect prediction stability.

We find statistical relationships between model hyperparameters and prediction stability.
However, it is not transparent how different aspects, such as model performance, model hyperparameters, and prediction stability, causally influence each other.
Attribution of changes to specific variables is difficult; hence, we only propose heuristics on how to select hyperparameters that minimize prediction instability.
However, as our experiments show, training a model with hyperparameters jointly selected according to these rules does decrease prediction instability.
Causal attribution and consequent robust rules for model selection with respect to prediction stability is another avenue for future work.

\noindent\textbf{Dataset Dependency. }
Repeatedly, models behave differently on the Pubmed dataset compared to the others.
This suggests that the dataset plays a crucial role in determining prediction stability.
We did not identify a single property of the dataset that sensibly explains the effect, which highlights the opportunity of examining prediction stability from the data perspective.

\noindent\textbf{Implementation as a Source of Instability. }
In additional experiments, we investigated the sources of instability by doing additional experiments with fixed random seeds and models trained on GPU vs CPU, see the supplementary material for full results.
We find that with a fixed random seed on the exact same data, GCN behaves completely deterministic while GAT in some cases still exhibits considerable instability.
We explain this by the fact that our implementation is based on Pytorch Geometric \cite{fey_fast_2019}, which uses nondeterministic scatter operations. Surprisingly, even when GAT is trained on a CPU with fixed initialization and training, minor instability remains.
Overall, GPU instabilities are much smaller compared to differences introduced by changing initialization.
This is good news from a pure reproducibility perspective, but we consider the instabilities established and analyzed in this paper still as crucial in many practical scenarios since they emphasize the sensitivity of predictions on implementation details and minuscule changes in the data of training and model.

\noindent\textbf{Influence on Model Selection. }
Our results have direct implications on model selection.
If we have to decide between multiple models that perform equivalently, and we are interested in minimizing prediction instability, then we can select the model with higher width and L2 regularisation, and lower dropout rate and depth.
While this may not be a straightforward decision, as tradeoffs between different variables have to be made, the proposed rule can be a rough guide.

\section{Related Work}
Our work is related to previous research that we outline in the following section.

\noindent\textbf{Stability of Node Embeddings. }
Wang et al. \cite{wang_towards_2020} and Schumacher et al. \cite{schumacher_effects_2020} study the influence of randomness on unsupervised node embeddings.
These embeddings, mainly computed via random walk-based models or matrix factorization, capture some notion of proximity of nodes, which should then be reflected in the geometry of the embedding space.
They both measure large variability in the geometry of the embedding spaces, e.g., in the nearest neighbors of embeddings.
Schumacher et al. find that the aggregated performance of downstream models does not change much, but individual predictions vary.
Wang et al. further demonstrate that less stable nodes are less likely to be predicted correctly.
In contrast to their work, we focus on supervised GNNs and prediction instability instead of geometrical instability.

\noindent\textbf{Impact of Tooling. }
Zhuang et al. \cite{zhuang_randomness_2021} find that prediction instability arises along the entire stack of software, algorithm design, and hardware.
Modifying model training to be perfectly reproducible incurs highly variable costs, in some cases more than tripling the computation time.
Moreover, they observe that subgroups of the data are affected to different extents from random factors in training.
Introducing batch normalization reduces performance variability but increases prediction disagreements.
They focus mainly on large CNNs, whereas we study comparatively much smaller GNNs.

\noindent\textbf{Increasing Prediction Stability. }
Several recent works successfully introduce techniques to increase prediction stability.
For example, by regularizing labels \cite{bahri_locally_2021,milani_fard_launch_2016}, distillation \cite{bhojanapalli_reproducibility_2021,jiang_churn_2021}, ensembling techniques \cite{shamir_anti-distillation_2020,summers_nondeterminism_2021}, or data augmentation \cite{summers_nondeterminism_2021}.
In some cases, it is noted that these techniques sometimes also increase model performance, but a general relationship between prediction stability and model performance is not highlighted.
For the CNNs in the work by Summers and Dineen \cite{summers_nondeterminism_2021}, even single bit changes lead to significantly different models.

\noindent\textbf{Model Influence. }
Liu et al. \cite{liu_model_2022} study how data updates affect prediction stability in the domain of language processing. 
Moreover, they compare whether model architecture, model complexity, or usage of pretrained word embeddings improve stability.
They identify a trade-off between prediction stability and model performance.
In our experiments, the trade-off is small or nonexistent, validating experiments of prediction stability in different domains.

\noindent\textbf{GNN Robustness. }
The stability of GNNs can be viewed from a different perspective: adversarial unnoticable perturbations of the graph data can significantly reduce model performance and thus prediction stability \cite{zugner_adversarial_2018}.
Zügner and Günnemann \cite{zugner_certifiable_2019} propose a method to certify robustness of nodes against such attacks.

\section{Conclusion}
In this paper, we systematically assessed the instability of Graph Neural Network predictions with respect to multiple aspects: random initialization, model architecture, data, and training setup.
We found that up to a third of the falsely predicted nodes are different between training runs that use the same data and hyperparameters but change the initialization.
Nodes on the periphery of a graph are less likely to be stably predicted.
Furthermore, models with higher width, higher L2 regularization, lower depth, and a lower dropout rate are more stable in their predictions.
Instability of deep GNNs is reflected in their internal representations.
Finally, maximizing model performance almost always implicitly minimizes prediction instability.

Future work may study prediction instability of GNNs from the perspective of larger, more complex models or data properties.
Furthermore, finding clear causal relationships may be beneficial to select models that are more stable with respect to their predictions.
Lastly, it would be interesting to see whether existing techniques aiming to reduce model instability for other types of models perform well for GNNs.

%
%
%
\bibliography{manual_refs.bib}
\end{document}